\pdfoutput=1
\documentclass[11pt]{article}

\usepackage[]{ACL2023}

\usepackage{times}
\usepackage{latexsym}
\usepackage{graphicx}
\usepackage{array}
\usepackage{booktabs}
\usepackage{graphicx, amsmath}
\usepackage{adjustbox}
\usepackage{caption}
\usepackage{multirow}
\usepackage{subcaption}
\usepackage{subcaption}     
\usepackage[T1]{fontenc}
\usepackage[utf8]{inputenc}

\usepackage{microtype}

\usepackage{inconsolata}

\usepackage{amsmath}
\usepackage[bb=px]{mathalfa} 
\title{\textit{Sem-DPO}: Mitigating Semantic Inconsistency in Preference Optimization for Prompt Engineering}

\author{
    \textbf{
    Anas Mohamed\textsuperscript{1*}\hspace{3em}
    Azal Ahmad Khan\textsuperscript{1*}
    } \\
    \textbf{
    Xinran Wang\textsuperscript{1}\hspace{3em}
    Ahmad Faraz Khan\textsuperscript{2}\hspace{3em}
    Shuwen Ge\textsuperscript{3}
    } \\
    \textbf{
    Saman Bahzad Khan\textsuperscript{4}\hspace{3em}
    Ayaan Ahmad\textsuperscript{5}\hspace{3em}
    Ali Anwar\textsuperscript{1}
    } \\
    \textsuperscript{1}University of Minnesota\hspace{2em} 
    \textsuperscript{2}Virginia Tech\hspace{2em}
    \textsuperscript{3}Xi'an University of Technology\\
    \textsuperscript{4}Lahore University of Management Sciences\hspace{2em}
    \textsuperscript{5}University of California, Santa Cruz\\
    \texttt{\{moha1325, khan1069, wang8740, aanwar\}@umn.edu}, \texttt{ahmadfk@vt.edu}\\
    \texttt{shuwen8681@gmail.com}, \texttt{27100111@lums.edu.pk}, \texttt{ayahmad@ucsc.edu}\\
}

\begin{document}
\maketitle

\begin{abstract}
Generative AI can now synthesize strikingly realistic images from text, yet output quality remains highly sensitive to how prompts are phrased.
Direct Preference Optimization (DPO) offers a lightweight, off-policy alternative to RL for automatic prompt engineering, but its token-level regularization leaves semantic inconsistency unchecked as prompts that win higher preference scores can still drift away from the user’s intended meaning.

We introduce Sem-DPO, a variant of DPO that preserves semantic consistency yet retains its simplicity and efficiency.
Sem-DPO adjusts the DPO loss using a weight based on how different the winning prompt is from the original, reducing the impact of training examples that are semantically misaligned.
We provide the first analytical bound on semantic drift for preference-tuned prompt generators, showing that Sem-DPO keeps learned prompts within a provably bounded neighborhood of the original text.
On three standard text-to-image prompt-optimization benchmarks and two language models, Sem-DPO achieves 8–12\% higher CLIP similarity and 5–9\% higher human-preference scores (HPSv2.1, PickScore) than DPO, while also outperforming state-of-the-art baselines.
These findings suggest that strong flat baselines augmented with semantic weighting should become the new standard for prompt-optimization studies and lay the groundwork for broader, semantics-aware preference optimization in language models.

\end{abstract}

\renewcommand{\thefootnote}{\fnsymbol{footnote}}
\footnotetext[0]{* Equal contributions (ordered via coin-flip).}
\renewcommand{\thefootnote}{\arabic{footnote}}

\section{Introduction}
\label{sec:Intro}

% \paragraph{Motivation} 
Recent advances in Generative AI have democratized creative expression, enabling users to generate high-quality images from textual input prompts~\citep{ramesh2021zero, saharia2022photorealistic, yu2022scaling, rombach2022high, ramesh2022hierarchical}. However, the quality of these outputs remains highly dependent on the design of input prompts, which must precisely represent styles, and contexts to guide generative models. Developing effective prompts is a non-trivial task, and users often resort to trial-and-error or manual engineering, a process that is labor-intensive and model-specific~\citep{reynolds2021prompt}.
As a result, recent work has investigated the use of Large Language Models (LLMs) to automate prompt engineering by paraphrasing and stylistic augmentation. This has led to the development of strategies for fine-tuning LLMs to generate prompts that optimize proxy metrics such as aesthetic score, using reinforcement learning (RL)-based approaches~\cite{hao2022optimizing, cao2023beautifulprompt}. However, these methods suffer from poor human alignment, since maximizing surrogate metrics does not necessarily reflect user preferences, as prompts optimized for aesthetic score, for instance, might produce overly stylized images that deviate from the user's intended meaning. Moreover, they can incur significant computational costs due to on-policy sampling (e.g., RLAIF)~\cite{lee2023rlaif, lindstrom2024ai, agarwal2024policy}.

%%%%%%%%%%%%%%%%%%%%%%%%%%%%%%%%
\begin{figure}[ht]
    \centering
    % Legend above the first two plots
    \begin{minipage}[b]{\linewidth}
        \centering
        \includegraphics[width=0.7\linewidth]{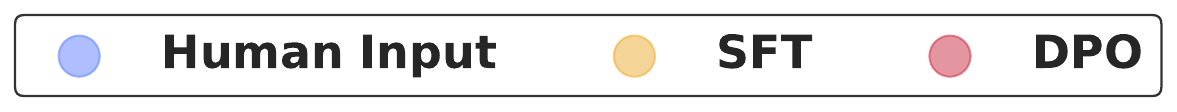}
    \end{minipage}
    \vspace{1em} 
    \begin{subfigure}[t]{0.4\textwidth}
        \centering
        \includegraphics[width=\linewidth]{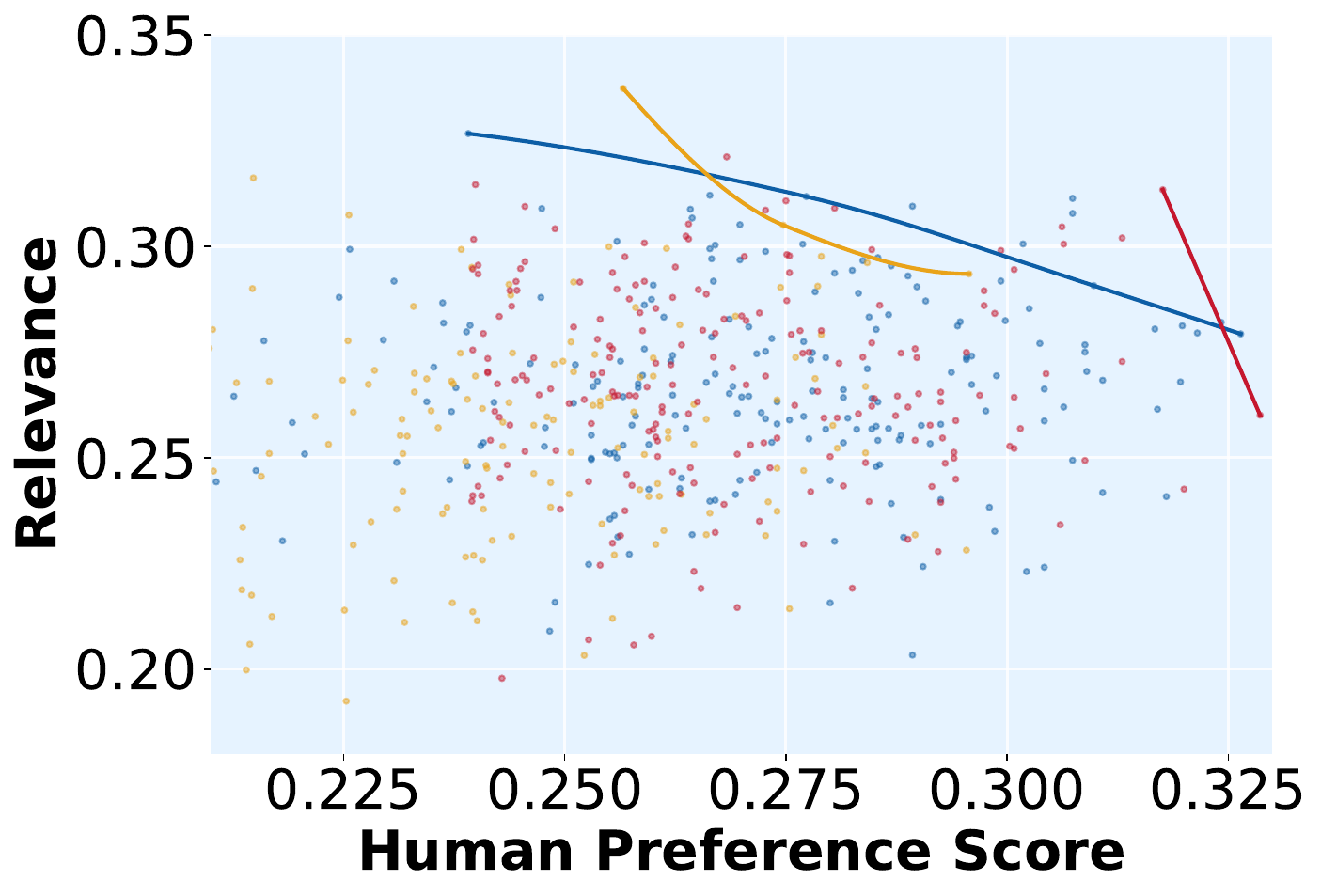}
        \label{fig:hpsvsclip}
    \end{subfigure}%
    \vspace{-1em}
    \caption{Comparison of prompt optimization methods for Human Preference metrics and semantic (CLIP) alignment. Legend shown above. SFT and DPO focus on maximizing HP but these methods lead to reducing semantic consistency.}
    \label{fig:all-together}
\end{figure}
%%%%%%%%%%%%%%%%%%%%%%%%%%%%%%%%

% \begin{figure*}
%     \centering
%     \includegraphics[width=\linewidth]{sections/Images/SDPO.pdf}
%     \caption{Comparing standard Direct Preference Optimization (DPO) with our Semantic-DPO variant. In standard DPO \textbf{(left)}, the policy LM is fine-tuned on all prompts equally, applying the same contrastive likelihood objective to every positive sample regardless of how much its meaning diverges from the original input. Semantic-DPO \textbf{(right)}, computes a semantic consistency weight based on the cosine distance between the input embedding ($e(x)$) and the preferred output embedding ($e(y_w)$). This weight is used to downscale the DPO loss for semantically distant samples, ensuring that fine-tuning preserves the original prompt's intended meaning.}
%     \label{fig:sdpo_main}
% \end{figure*}

\paragraph{Challenges.}
To address the challenges of human alignment and cost of RL-based approaches, Direct Preference Optimization (DPO) has emerged as a compelling off-policy alternative~\cite{rafailov2023direct}. DPO aligns the policy with preference data by maximizing a contrastive likelihood objective over preferred and dispreferred samples, typically derived from human or proxy annotator comparisons. While DPO has shown strong empirical performance in dialogue and instruction tuning tasks~\cite{saeidi2024insights, dong2024self, jung2025diatool}, its application to prompt optimization reveals a critical weakness.
As illustrated in Figure~\ref{fig:all-together}, DPO improves human preference alignment, i.e., it generates outputs more liked by humans, but often reduces semantic consistency, where outputs stray from the original prompt’s meaning (as measured by CLIP). This suggests that the DPO-optimized prompts, though more aligned with human preferences, can drift semantically from the intended meaning. We attribute this to the structure of the DPO gradient, which is driven by token-level likelihood ratios and provides no signal for preserving the semantic distance between the input and the preferred output~\cite{razin2024unintentional, yang2024orthogonal}.
Recent work shows that even sophisticated prompt optimizers \citep{zhu2025rethinking} and pair-reweighted DPO variants \citep{amini2024direct} still overlook semantic drift.

% This findings motivate the need for a preference optimization framework that is sensitive not only to human preference but also to semantic consistency.

\paragraph{Prior Works.}
Several variants of DPO have been proposed to improve optimization stability and alignment robustness, such as KL-regularized DPO~\cite{wang2023beyond}, T-REG~\cite{zhou2024t}, $\beta$-DPO~\cite{wu2024beta} and two-stage filtering pipelines~\cite{shan2025forward}. 
More recently, semantic-aware Kernel-based Preference Optimization (KTO;~\citet{ethayarajh2024kto}) have been introduced.
KTO compares model outputs using divergence-based similarity metrics, enabling the model to capture richer preference structures beyond simple log-likelihood ratios and helping preserve semantic content during training. However, these approaches still primarily operate within the token-probability or distributional space, lacking direct integration of semantic meaning into the loss function itself. Critically, they do not penalize preference pairs in which the selected output is semantically misaligned with the input, thereby implicitly rewarding stylistic overfitting. 

This motivates our central question. 
{\textit{How can we adapt DPO to optimize prompts so the outputs are both highly preferred by humans and remain semantically faithful to the original input?}}

\paragraph{Our Approach.}
We propose \emph{Sem-DPO} (Semantic-DPO), a variant of DPO that explicitly incorporates semantic consistency into the preference alignment process. While current methods focus primarily on optimizing token-level log-likelihood ratios, Sem-DPO introduces an importance weighting mechanism that adjusts the influence of each training sample based on its semantic alignment with the original prompt. Specifically, for each triplet $(x, y_w, y_l)$, we compute a semantic consistency weight $W(x, y_w) = \exp(-\alpha \cdot d(e_{\varphi}(x), e_{\varphi}(y_w)))$, where $e_{\varphi}(\cdot)$ is an embedding function and $d(\cdot, \cdot)$ is a cosine distance metric. This weight downscales the gradient contribution of preference pairs where the preferred output semantically diverges from the input, thereby mitigating the semantic inconsistency often observed in prompt optimization. Crucially, the weighting is computed offline and integrated to the loss function, preserving the computational efficiency and training simplicity of DPO. Sem-DPO thus fills the gap of DPO framework to enforce preference alignment and semantic consistency, making it better for tasks where preserving input intent is critical.

\paragraph{Contributions.}
This work makes \emph{four decisive advances} toward trustworthy prompt optimization for text-to-image models.
\textbf{(C1) Problem Formulation.}  We are the first to cast \emph{semantic drift} as a core failure mode of preference-based prompt optimization, turning an often-ignored side-effect into a formal learning objective.
\textbf{(C2) Semantically-Weighted DPO Objective.}  We introduce \emph{Sem-DPO}, a lightweight yet powerful variant of DPO that injects semantic-aware importance weights into the loss, seamlessly combining human preference alignment with semantic fidelity, \emph{without} sacrificing DPO's performance and simplicity.
\textbf{(C3) Theoretical Analysis.}  We provide the \emph{first} theoretical guarantees that explicitly bound semantic drift under preference optimization, offering provable insight into when and why Sem-DPO outperforms its prior works.
\textbf{(C4) Experimental Analysis.}  Extensive experiments on three benchmark datasets and two language models show that Sem-DPO consistently delivers the best trade-off between human preference and semantic alignment, establishing a new state of the art in prompt optimization.

% This paper makes the following key contributions:
% \textbf{(C1)} We formalize the problem of semantic inconsistency in the context of preference-based prompt optimization.
% \textbf{(C2)} We propose Sem-DPO, a semantically-aware variant of DPO that incorporates meaning-level alignment during preference optimization.
% \textbf{(C3)} We supply the theoretical analysis of preference-based prompt optimization with semantic constraints.
% \textbf{(C4)}  We evaluate Sem-DPO on several text-to-image prompt optimization benchmarks.

\section{Related Works}
\label{sec:Related}

\paragraph{Preference Alignment.}
Early alignment research framed the problem as RLHF, where a reward model trained on pairwise human preferences guides policy optimization with PPO or variants~\cite{bai2022training, ouyang2022training}.
Although effective, RLHF imposes a heavy engineering overhead and a large on-policy sampling cost. DPO eliminates the explicit reward model by casting alignment as a contrastive classification problem whose optimal policy is available in closed form~\cite{rafailov2023direct}, matching or exceeding RLHF while being simpler and compute-efficient.
DPO has demonstrated strong empirical results in dialogue and instruction tuning tasks, and its simplicity and efficiency have led to its adoption in various domains~\cite{dong2024self, yang2024enhancing, sun2024conifer, lu2024fipo, khaki2024rs}.
Since then, DPO has been developed for multiple specialized axes like domain generalization~\cite{wallace2024diffusion}, contrastive refinements~\cite{xiao2024cal, omura2024entropy}, and theoretical analyses~\cite{park2024disentangling}.

\paragraph{Regularization.} 
Recent work introduces explicit regularizers to the DPO loss to restrain over-confident or reward-hacking behaviours.
Entropy-based penalties adjust the exploration–exploitation balance during fine-tuning~\cite{omura2024entropy}, while calibrated objectives match the scale of ground-truth rewards~\cite{xiao2024cal}. Feature-level constraints restrict updates to sparse or disentangled subspaces to preserve fluency~\cite{yin2024direct}, and game-theoretic self-play adds adversarial robustness~\cite{tang2025game}.
Recent efforts edge toward semantic regularization, \emph{DPO-Kernels} couples probability-space losses with embedding-space kernels~\cite{das2025dpo}, \emph{DSPO} injects instance-level semantic guidance for real-image super-resolution~\cite{cai2025dspo}, and Hallucination-Aware DPO penalises vision-language mismatch~\cite{zhao2023beyond}. Many existing regularization techniques are tailored to specific modalities or require additional annotation, limiting their scalability and general applicability~\cite{shekhar2024see, jinnai2024annotation}.
Further re-weighting schemes such as Weighted-PO adjust gradients according to preference margins alone, leaving semantic drift unaddressed \citep{zhou2024wpo}.
\emph{Therefore, existing regularizers either treat semantics as a post-hoc filter or rely on ad-hoc domain heuristics and none integrate semantic consistency directly into the core preference loss or offer guarantees against prompt-level drift.}

\newtheorem{proposition}{Proposition}
\newtheorem{definition}{Definition}
\newtheorem{assumption}{Assumption}
\newtheorem{corollary}{Corollary}
\newcommand*{\medcap}{\mathbin{\scalebox{1.5}{\ensuremath{\cap}}}}

\section{Proposed Approach}
This section begins by providing a theoretical background of DPO. Following this, we introduce Sem-DPO, our proposed variant of DPO that explicitly incorporates semantic consistency while retaining DPO's simplicity and efficiency.

\paragraph{Preliminaries.}
Reinforcement Learning from Human Feedback (RLHF) aligns models with human preferences using a dataset of triplets $\mathcal{D} = \{(x, y_w, y_l)\}$, where a response $y_w$ is preferred over $y_l$ for a given input $x$. While traditional RLHF first trains a reward model and then uses it to optimize a policy, DPO streamlines this into a single-stage training process. DPO achieves this by establishing a direct analytical link between the reward function and the optimal policy $\pi^*$. Specifically, the reward function is reparameterized in terms of the policy as:

{\small
\[
r^* (x,y) = \beta \log \frac{\pi^* (y|x)}{\pi_{\text{ref}} (y|x)} + \beta \log Z(x)
\]
}

By substituting this relationship into the Bradley-Terry preference model, DPO derives a simple maximum likelihood objective that directly optimizes the language model policy $\pi_\theta$ on the preference data. This avoids the complexities of training a separate reward model by instead minimizing the following negative log-likelihood loss:

{\small
\begin{equation}
\begin{split}
\mathcal{L}_{\text{DPO}}(\pi_\theta)= -\mathbb{E}_{(x, y_w, y_l) \sim \mathcal{D}} \\
\left[ \log \sigma\left( \beta \log \frac{\pi_\theta(y_w|x)}{\pi_{\text{ref}}(y_w|x)} - \beta \log \frac{\pi_\theta(y_l|x)}{\pi_{\text{ref}}(y_l|x)} \right) \right]
\end{split}
\label{eq:DPOloss}
\end{equation}
}

\begin{figure}
    \centering
    \includegraphics[width=\linewidth]{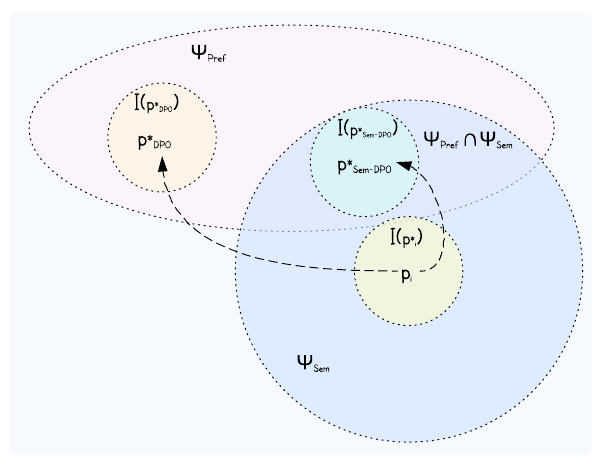}
    \vspace{-2em}
    \caption{\textbf{Concept Figure}. DPO steers prompts into the preference region, while the ideal target lies at the intersection of preference and semantic regions. Sem-DPO is designed to reach that overlap}
    \label{fig:concept}
\end{figure}

\subsection{Motivation}
The goal of automated prompt optimization is to refine an initial user prompt, $p_\text{i}$, to generate a superior image. An ideal prompt must satisfy two conditions: it must produce a high-quality image preferred by humans, thus lying in a high-preference region ($\Psi_{\text{Pref}}$), and it must preserve the core meaning of the original request, remaining within a semantic consistency space ($\Psi_{\text{Sem}}$), as shown in Figure~\ref{fig:concept}. While powerful methods like DPO are effective at navigating the prompt space to find solutions ($p^*_{\text{DPO}}$) that score high on preference metrics, they suffer from a critical flaw. Since DPO operates at a token level without an intrinsic understanding of meaning, its optimization path often exits the semantic consistency space. This ``Semantic Drift'' results in a final prompt that, although optimized for preference, generates an image that is semantically inconsistent from the user's intent.

To address this, we propose Sem-DPO, a method designed to find an optimal prompt, $p^*_{\text{Sem-DPO}}$, that resides in the intersection of both the preference and semantic spaces ($\Psi_{\text{Pref}} \cap \Psi_{\text{Sem}}$). Our approach augments the standard DPO objective with a semantic-aware weighting mechanism. This mechanism calculates the semantic distance between candidate prompts and the initial prompt $p_\text{i}$, penalizing any updates that cause significant semantic drift. This penalty acts as a corrective force, constraining the optimization search to remain within the semantic consistency space while simultaneously seeking higher-preference solutions. As a result, Sem-DPO produces a prompt that is not only preferred but also semantically faithful, ensuring the final generated image successfully aligns with the user's intent.

\paragraph{Problem Formulation}
The task of preference-based prompt optimization is to fine-tune a policy model, $\pi_{\theta}$ to generate improved prompts. Standard DPO achieves this by minimizing Equation~\ref{eq:DPOloss}.

\begin{definition}[Semantic Drift]
\label{prop:SemanticDrift}
Let \(e_{\varphi}:\mathcal{X}\!\to\!\mathbb{R}^{\delta}\) be a frozen pre-trained embedding model and \(d_{\text{cos}}(\cdot,\!\cdot)\) a cosine distance measure.  
For an initial prompt \(x\) and an optimized prompt \(y\), we define the \emph{semantic drift} $d(x,y)=d_{\text{cos}}\bigl(e_{\varphi}(x),\,e_{\varphi}(y)\bigr)$.
The prompt y exhibits significant \emph{semantic drift} whenever \(d(x,y)\ge \tau\) for a user-chosen threshold.
\end{definition}

The central issue we address is that the DPO loss operates at the token level and includes no term to penalize semantic drift. Consequently, a DPO-finetuned model may generate a prompt that produces a stylistically superior yet semantically inconsistent image with the original prompt, a phenomenon we observe empirically (Figure~\ref{fig:all-together}).

% \textbf{Definition 1:} \textit{We define semantic drift as the semantic divergence between an initial prompt ($x$) and a refined prompt ($y$). This drift can be quantified by a distance metric $d(e_{\varphi}(x), e_{\varphi}(y_{w}))$, where $e_{\varphi}(\cdot)$ is a pre-trained, frozen embedding model, and $d(\cdot, \cdot)$ is a distance function. A prompt $y_w$ exhibits significant semantic drift if $d(e_{\varphi}(x), e_{\varphi}(y_{w})) \ge \tau$ for some threshold $\tau$.}

\subsection{\textit{Sem-DPO}: Semantic Direct Preference Optimization}
To mitigate the semantic drift in the standard DPO framework, we propose Semantic Direct Preference Optimization (Sem-DPO). Our approach modifies the DPO objective by introducing a per-sample semantic weight that is a function of the semantic similarity between the input prompt ($x$) and the preferred output ($y_w$).
Specifically, we define a semantic consistency weight $W_{\alpha}(x,y_w)$ as:

\begingroup
\small
\begin{equation}
W_{\alpha}(x, y_w) = \exp\biggl(-\alpha \cdot d_{\text{cos}}\Bigl(e_{\varphi}(x), e_{\varphi}(y_w) \Bigr) \biggr)
\label{eq:weighting}
\end{equation}
\endgroup

where, $\alpha \ge 0$ is a hyperparameter that controls the strength of the semantic weighting. This weight is then incorporated directly into the DPO loss function, yielding the Sem-DPO objective:

{
\small
\[
\begin{split}
\mathcal{L}_{\text{Sem-DPO}}(\theta) = -\mathbb{E}_{(x, y_w, y_l) \sim \mathcal{D}} \\
W_{\alpha}(x, y_w) \cdot \log \Biggl( \sigma \biggl( \beta \Bigl( 
\log \frac{\pi_\theta(y_w|x)}{\pi_{ref}(y_w|x)} - \frac{\pi_\theta(y_l|x)}{\pi_{ref}(y_l|x)} \Bigr) \biggr) \Biggr)
\end{split}
\]
}

This weighting scheme (Equation~\ref{eq:weighting}) down-weights training samples where the preferred prompt semantically drifts from the input. Pairs with high semantic similarity receive higher weights, preserving their gradient influence, while those with low similarity are exponentially suppressed. This encourages outputs that are both aesthetically and semantically aligned, ensuring the optimized prompt stays within the target region $\Psi_{\text{Pref}} \cap \Psi_{\text{Sem}}$. Since weights are computed offline using a frozen encoder, this incurs no additional overhead and retains the original DPO's efficiency.
\section{Theoretical Analysis}
\label{sec:theory}
We present a theoretical analysis of Sem-DPO, establishing two key results that justify its design. First, we show that the exponential weighting provides a smooth, stable approximation to hard semantic filtering. Second, we prove that by controlling semantic drift, Sem-DPO bounds the drift between the generated image and the input prompt. The log-odds difference used in DPO is:

\begingroup
\small
\begin{equation}
  \Delta_\theta(x,y_w,y_l) = \beta \left( \log \frac{\pi_\theta(y_w \mid x)}{\pi_{\text{ref}}(y_w \mid x)} - \log \frac{\pi_\theta(y_l \mid x)}{\pi_{\text{ref}}(y_l \mid x)} \right)
\end{equation}
\endgroup

\begingroup
\small
\begin{equation}
  \ell(\Delta) = -\log \sigma(\Delta)
\end{equation}
\endgroup

\paragraph{Smooth Filtering Mechanism.}
Sem-DPO employs a smooth exponential kernel to down-weight semantically divergent samples, avoiding the discontinuities introduced by hard filtering based on a fixed threshold. Our first proposition shows that this yields a stable relaxation of hard filtering, with bounded approximation error.

\begin{proposition}[Uniform Deviation Bound]
\label{prop:uniform_bound}
Sem-DPO's exponential importance-weighting, $W_{\alpha}(d) = \exp(-\alpha d)$, serves as a smooth relaxation of hard semantic filtering, with a bounded approximation error.
\end{proposition}

\emph{Proof.} Let hard semantic filtering be defined by an objective function that uses an indicator function $1_{\{d \le \tau\}}$ to discard examples where the semantic distance $d$ exceeds a threshold $\tau > 0$:
\[
    \mathcal{L}_{\tau}(\theta) = \mathbb{E}_{\mathcal{D}}[1_{\{d \le \tau\}} l(\Delta_{\theta})]
\]
where $l(\Delta_{\theta}) = -\log \sigma(\Delta_{\theta})$ is the standard DPO loss term.

Sem-DPO replaces this discontinuous indicator function with the smooth exponential kernel $W_{\alpha}(d)$, yielding the objective:
\[
    \mathcal{L}_{\text{Sem-DPO}}(\theta) = \mathbb{E}_{\mathcal{D}}[W_{\alpha}(d) l(\Delta_{\theta})]
\]

Assuming the loss term is bounded, such that $|l(\cdot)| \le M$ for some constant $M$, the absolute deviation between the Sem-DPO objective and the hard-filtered objective is bounded as follows:
{\small
\[
\begin{aligned}
\bigl|\mathcal{L}_{\text{Sem-DPO}}(\theta)-\mathcal{L}_{\tau}(\theta)\bigr|
&=\Bigl|\mathbb{E}_{\mathcal D}\!\bigl[(W_\alpha(d)-\mathbf 1_{d\le \tau})\,\ell(\Delta_\theta)\bigr]\Bigr|\\
&\le\mathbb{E}_{\mathcal D}\!\bigl[\bigl|W_\alpha(d)-\mathbf 1_{d\le \tau}\bigr|\,\bigl|\ell(\Delta_\theta)\bigr|\bigr]\\
&\le M\,\mathbb{E}_{\mathcal D}\!\bigl[\bigl|W_\alpha(d)-\mathbf 1_{d\le \tau}\bigr|\bigr].
\end{aligned}
\]
}

The maximum pointwise difference between the weighting function $W_{\alpha}(d)$ and the indicator function occurs at the discontinuity point $d=\tau$ and is bounded by $1 - e^{-\alpha\tau}$. Thus, the expected deviation is also bounded:
\[
    |\mathcal{L}_{\text{Sem-DPO}}(\theta) - \mathcal{L}_{\tau}(\theta)| \le M(1 - e^{-\alpha\tau})
\]
which completes the proof.

\emph{Interpretation.} Proposition~\ref{prop:uniform_bound} justifies the exponential weighting as a stable approximation to hard semantic filtering. It avoids gradient discontinuities while allowing control over filter sharpness via the hyperparameter $\alpha$, enabling robust enforcement of semantic consistency.

\paragraph{Guarantee for Semantic Consistency.}
We now present our main theoretical result, which provides a formal guarantee that Sem-DPO achieves its ultimate goal of ensuring the final, high-preference image is semantically aligned with the original user prompt. This is predicated on a reasonable assumption about the quality of the underlying text-to-image generator.

\begin{assumption}[T2I Consistency Error]
\label{assump:T2IConsistency}
For any prompt $y$ and a high-quality text-to-image generator $G$, the consistency error $d_{\text{T2I}}(y) = \|e_(y) - E_{\text{Img}}((G(y))\|$, which measures the distance between the prompt's embedding and its corresponding image's embedding in a shared space, is bounded by a small constant $\epsilon$.
\end{assumption}
\[
    \|e_{(y)} - E_{\text{Img}}(G(y))\| \le \epsilon
\]
This implies the generated image remains semantically close to its prompt, within an $\epsilon$-bounded neighborhood in embedding space.

\begin{proposition}[Bounded Semantic Drift]
\label{prop:cross_modal_bound}
The semantic distance between the original prompt $x$, optimized prompt $y$ and the final image $G(y)$, denoted $d_{\text{T2I-Drift}}(x, y)$, is upper-bounded by the sum of the semantic drift and the T2I consistency error of the generator.
\end{proposition}

\emph{Proof.}
Let $e(x)$ and $e(y)$ be the text embeddings of prompts $x$ and $y$, and let $E_{\text{Img}}(G(y))$ be the image embedding of the generated image from prompt $y$. The text-to-image drift is defined as $d_{\text{T2I-Drift}}(x, y) = \| e_(x) - E_{\text{Img}}(G(y)) \|$.

By applying the triangle inequality property of vector norms, we can establish an upper bound:

{
\small
\[
\begin{split}
\| e(x) - E_{\text{Img}}(G(y)) \| \le \\
\| e(x) - e(y) \| + \| e(y) - E_{\text{Img}}(G(y)) \|
\end{split}
\]
}

This inequality can be expressed using our distance notations:

\[
    d_{\text{T2I-Drift}}(x, y) \le d_{\text{Semantic-Drift}}(x, y) + d_{\text{T2I}}(y)
\]
Applying our T2I Consistency Error Assumption, where $d_{\text{T2I}}(y) \le \epsilon$, we arrive at the final bound:
\[
    d_{\text{T2I-Drift}}(x, y) \le d_{\text{Semantic-Drift}}(x, y) + \epsilon
\]
which completes the proof.

\emph{Interpretation.} Proposition~\ref{prop:cross_modal_bound} justifies Sem-DPO by showing that the total semantic drift between the original prompt and the final image is bounded by the sum of a controllable semantic drift and a small, fixed generator error $\epsilon$. Sem-DPO explicitly minimizes the prompt drift through semantic weighting, ensuring alignment within the limits of the generator. This decomposition highlights that while $\epsilon$ is irreducible, the key to improving semantic consistency lies in controlling what is optimizable.
%%%%%%%%%%%%%%%%%%%%%%%%%%%%%%%%
\begin{figure*}[ht] 
    \centering 

    % legend on top centered
    \includegraphics[width=0.8\textwidth]{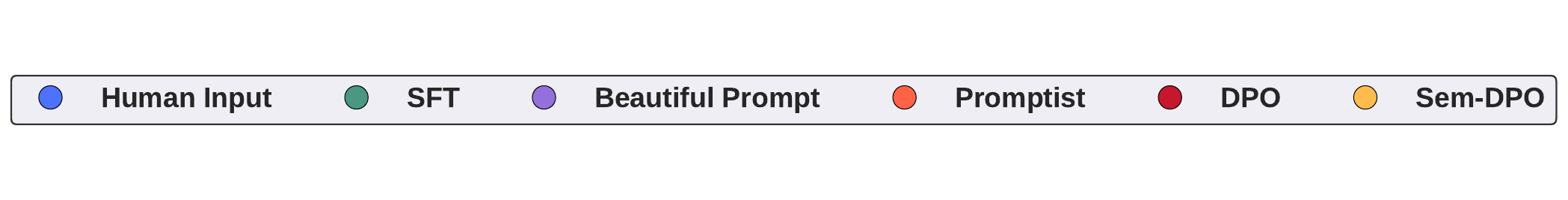}
    \vspace{-0.5em}

    % Subfigures for the two plots side-by-side
    \begin{subfigure}{\textwidth} % Adjust width as needed
        \centering
        \includegraphics[width=\linewidth]{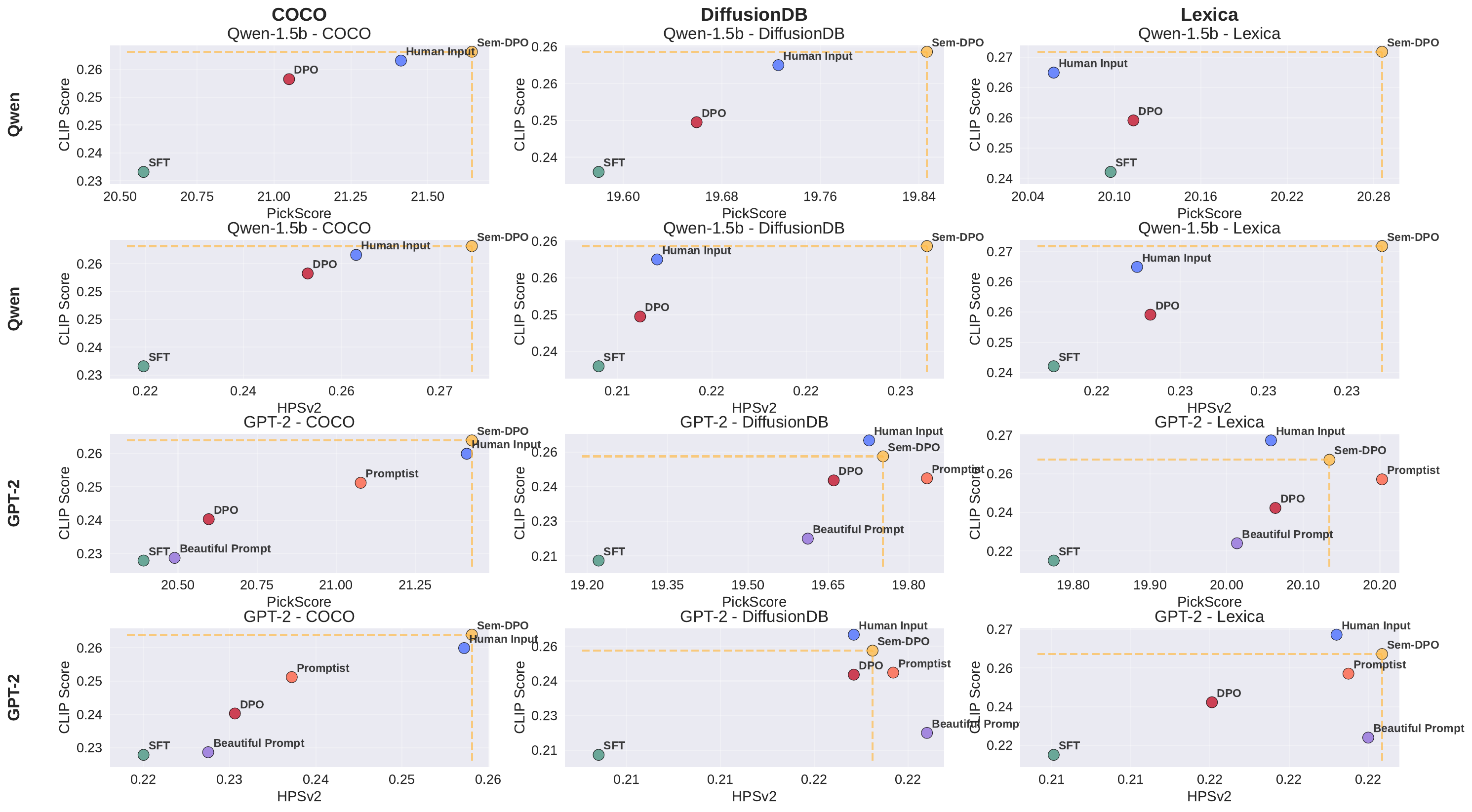}
        % \caption{CLIP Score vs. HPS v2.1 Score} 
        \label{fig:clip_vs_hps}
    \end{subfigure}

    % \vspace{-1em} 
    \caption{\textbf{Semantic–preference landscape of prompt-optimization methods across models and datasets.}
Scatter plots chart CLIP Score against human preference (PickScore and HPSv2.1) on three datasets with Qwen-1.5b (top) and GPT-2 (bottom) generators. Points nearer the upper-right indicate prompts excelling in both metrics.}
    \label{fig:combined_performance_plots}
\end{figure*}
%%%%%%%%%%%%%%%%%%%%%%%%%%%%%%%%

\section{Evaluation}
\label{sec:Exp}
In this section, we present the experimental validation of Sem-DPO. We detail the setup, including datasets, baselines, and metrics, and then discuss the comprehensive results comparing Sem-DPO against other prompt optimization approaches.

\subsection{Experimental Setup}
This section details the experimental setup used to evaluate our proposed Sem-DPO method, including the datasets, baselines, evaluation metrics, and implementation specifics.

\paragraph{Datasets.}
To assess the efficacy of Sem-DPO, we conduct experiments on three publicly available datasets. For in-domain testing, we utilize DiffusionDB and Lexica. For out-of-domain generalization assessment, we use the COCO dataset. Following prior work, the prompts for our experiments are sourced from ~\citet{hao2022optimizing}, with 200 prompts selected from each dataset. 

\paragraph{Baselines.}
We evaluate \textsc{Sem-DPO} under two prompt–generation backbones to provide both a strong contemporary comparison and continuity with prior work. Our primary setting uses \textbf{Qwen-1.5B} as the prompt generator, here we compare against the original Human Inputs, Supervised Fine-Tuning (SFT) variant that fine-tunes Qwen-1.5B with supervised losses, and DPO variant that fine-tunes the same backbone without our semantic term. \textit{To align with earlier prompt-optimization studies that build on older models (GPT-2), we additionally repeat all experiments with a GPT-2 backbone} and, in that setting, report results for \textbf{Promptist}~\cite{hao2022optimizing} and \textbf{BeautifulPrompt}~\cite{cao2023beautifulprompt} alongside GPT-2 versions of SFT and DPO. This dual-backbone design yields a fair, apples-to-apples comparison with prior literature while highlighting the incremental gains delivered by \textbf{Sem-DPO} on newer models.

\paragraph{Metrics.}
We evaluate the generated images using a combination of metrics to assess semantic consistency and human preference alignment. These include the CLIP Score, to measure the semantic relevance between the generated image and the input prompt, where higher scores indicate better semantic alignment. We also use PickScore~\cite{kirstain2024pick}, an image quality metric trained on human preferences reflecting the likelihood a human would prefer a given image, and HPSv2.1~\cite{wu2023human}, a Human Preference Score also trained to predict human judgments of image quality and alignment. These metrics provide quantitative measures to assess the quality and relevance of the generated images based on different criteria.

\paragraph{Preference Labels and Finetuning.} For 50k inputs, candidate prompts were generated with a SFT model. Each candidate was then rendered into an image using Stable Difussion v1.4 with the DDIM scheduler. The images were then scored by our preference model (ImageReward~\cite{xu2024imagereward}) and were assigned “chosen” or “rejected” labels which served as the training signal for both DPO and Sem-DPO. Completing this end to end preference generation workflow required approximately 26 hours on a single NVIDIA RTX 4090 GPU. 
Model fine tuning (both DPO and Sem-DPO) was carried out on a single NVIDIA A100 GPU. Training the DPO baseline required approximately 1 hour. Sem-DPO performed similarly, with about a ten minute increase in training time comparatively. 

\paragraph{Experimental Details.} Our experiments utilize the Stable Diffusion v1.4 text-to-image model. To expedite the image sampling process, we employ the Denoising Diffusion Implicit Models scheduler\citep{song2020denoising}. The number of denoising steps was set to 20 for all image generation tasks.  For both DPO and Sem-DPO, the $\beta$ parameter was set to 0.05. For Sem-DPO, the semantic weighting parameter $\alpha$ was set to 8 after manual tuning.

\subsection{Experimental Results}

\paragraph{Performance and Semantic Consistency.}
The efficacy of Sem-DPO was rigorously evaluated against standard and state-of-the-art prompt optimization methods, leveraging both automated alignment metrics (CLIP Score) and human preference metrics (HPSv2.1 and PickScore). As shown in Figure~\ref{fig:combined_performance_plots} and Table~\ref{table:qwen_results} \& \ref{table:gpt_results} in Appendix~\ref{sec:appendix}, Sem-DPO consistently demonstrated superior performance, particularly in enhancing human preference alignment while maintaining strong semantic consistency across diverse datasets, thereby outperforming previous baselines of prompt optimization in text-to-image generation. As depicted in Figure~\ref{fig:all-together}, standard DPO often improved human alignment over Original Prompt but frequently resulted in lower CLIP scores, indicative of semantic inconsistency where generated prompts diverged from the user's original intent. Sem-DPO addresses this limitation by explicitly incorporating semantic consistency into the optimization process.

\paragraph{Analysis Across Datasets.}
A detailed analysis of the results, as presented in Table~\ref{table:qwen_results} \& \ref{table:gpt_results} (Appendix~\ref{sec:appendix}), underscores Sem-DPO's advancements across in-domain (DiffusionDB \& Lexica) and out-of-domain (COCO) datasets. For DiffusionDB, Sem-DPO achieved a better HPSv2.1 score and PickScore compared to other baselines in most of the scenarios, while maintaining a strong CLIP Score. On the Lexica dataset, Sem-DPO exhibited the highest CLIP Score, HPSv2.1 score, and PickScore among all evaluated methods, highlighting its robustness in an in-domain setting. In out-of-domain, Sem-DPO continued to excel, yielding the highest HPSv2.1 score and a PickScore, further affirming its generalization capabilities and superior human preference alignment. Although the absolute gains in PickScore appear modest, they translate into noticeably improved visual fidelity in the generated images, also shown in prior works.

%%%%%%%%%%%%%%%%%%%%%%%%%%%%%%%%
\begin{figure*}[ht]
    \centering
    % Legend above the first two plots
    \begin{minipage}[b]{\linewidth}
        \centering
        \includegraphics[width=0.5\linewidth]{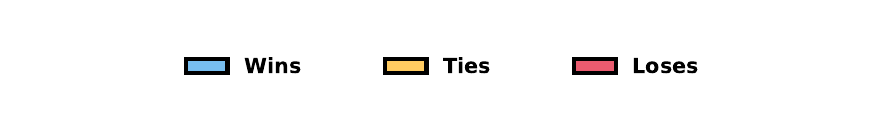}
    \end{minipage}

    % Three subfigures
    \begin{subfigure}[t]{0.33\textwidth}
        \centering
        \includegraphics[width=\linewidth]{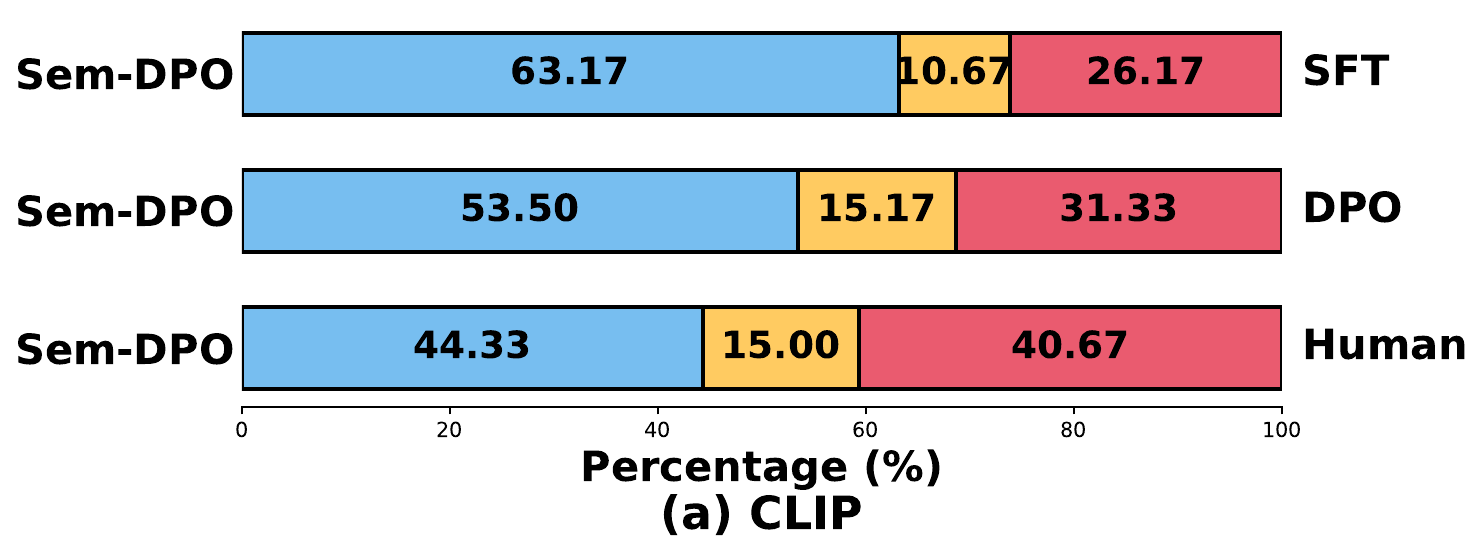}
        \label{fig:plot_clip_comparison}
    \end{subfigure}%
    \begin{subfigure}[t]{0.33\textwidth}
        \centering
        \includegraphics[width=\linewidth]{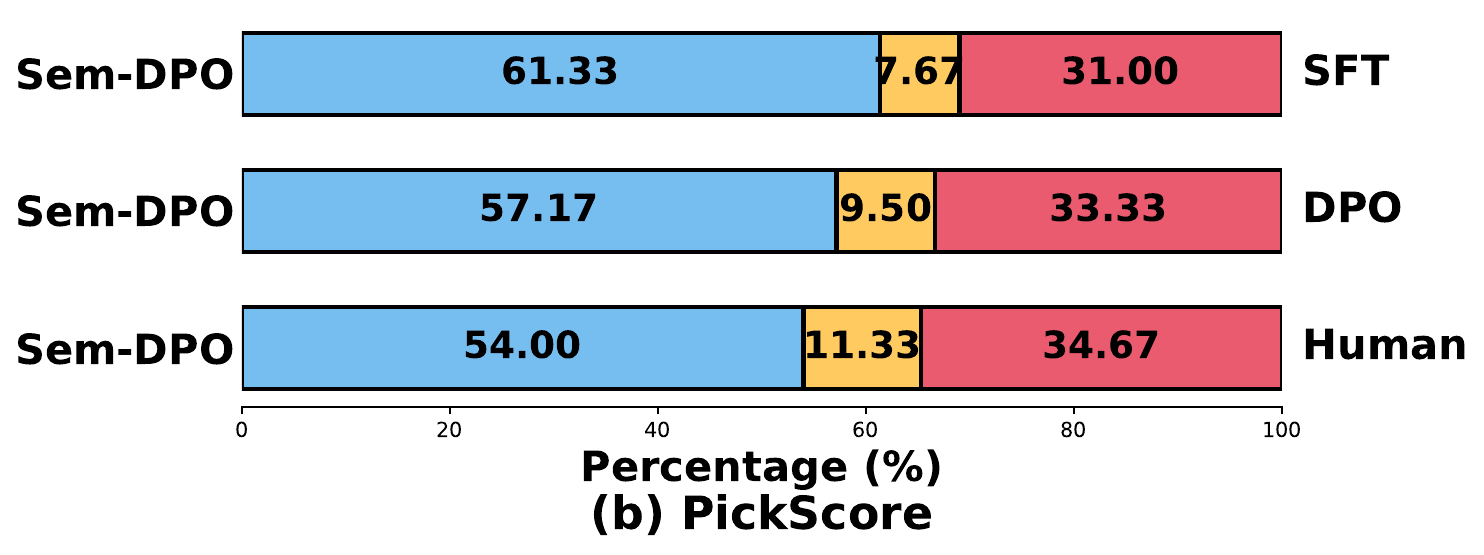}
        \label{fig:plot_pickscore_comparison}
    \end{subfigure}%
    \begin{subfigure}[t]{0.33\textwidth}
        \centering
        \includegraphics[width=\linewidth]{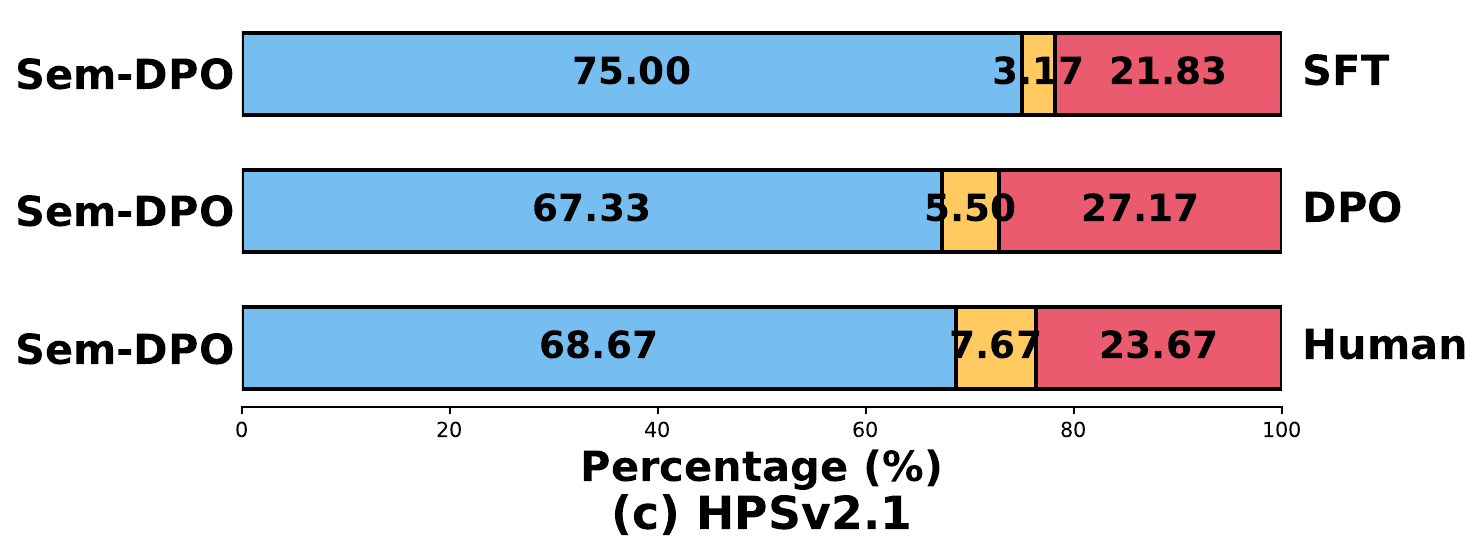}
        \label{fig:plot_hpsv2_comparison}
    \end{subfigure}%
    \vspace{-0.5em}
    \caption{Head-to-head comparison of Sem-DPO against three baselines (SFT, Base DPO, and Human Input) across three evaluation metrics.}
    \label{fig:comparison}
\end{figure*}
%%%%%%%%%%%%%%%%%%%%%%%%%%%%%%%%

\paragraph{Performance Insights.}
Figure~\ref{fig:combined_performance_plots} further confirms these findings. Specifically, the plots show CLIP score against HPSv2.1 score and CLIP score against PickScore, collectively demonstrate that Sem-DPO points are generally positioned towards the upper-right region. This placement signifies a desirable balance where Sem-DPO simultaneously achieves higher human preference scores and maintains competitive semantic alignment (CLIP Score) compared to other approaches like SFT, Promptist, BeautifulPrompt, and standard DPO. These experimental outcomes collectively validate that Sem-DPO not only effectively mitigates semantic inconsistency but also substantially improves alignment with human preferences, advocating for semantics-aware weighting as a new standard in prompt optimization studies.

Figure~\ref{fig:comparison} reports how often a Sem-DPO-generated prompt wins, ties, or loses when pitted one-by-one against outputs from Human Input, base DPO, and SFT under each metric. Against DPO, Sem-DPO wins on 53.5\% of prompts for CLIP, 57.2\% for PickScore, and 67.3\% for HPS v2.1, with losses dropping to roughly one-third of the cases.
The margin is even larger over SFT (e.g., 63.2\% wins on CLIP and 75.0\% on HPS), while gains over raw Human Input are positive but smaller, reflecting that Sem-DPO occasionally prefers stylistic tweaks that do not always boost semantic alignment.
Overall, the chart shows a clear majority-win pattern, confirming that Sem-DPO is consistently preferred to both implementation baselines and the original human prompts across all three evaluation axes.

%%%%%%%%%%%%%%%%%%%%%%%
\begin{figure}[ht]
    \begin{subfigure}[t]{0.24\textwidth}
        \centering
        \includegraphics[width=\linewidth]{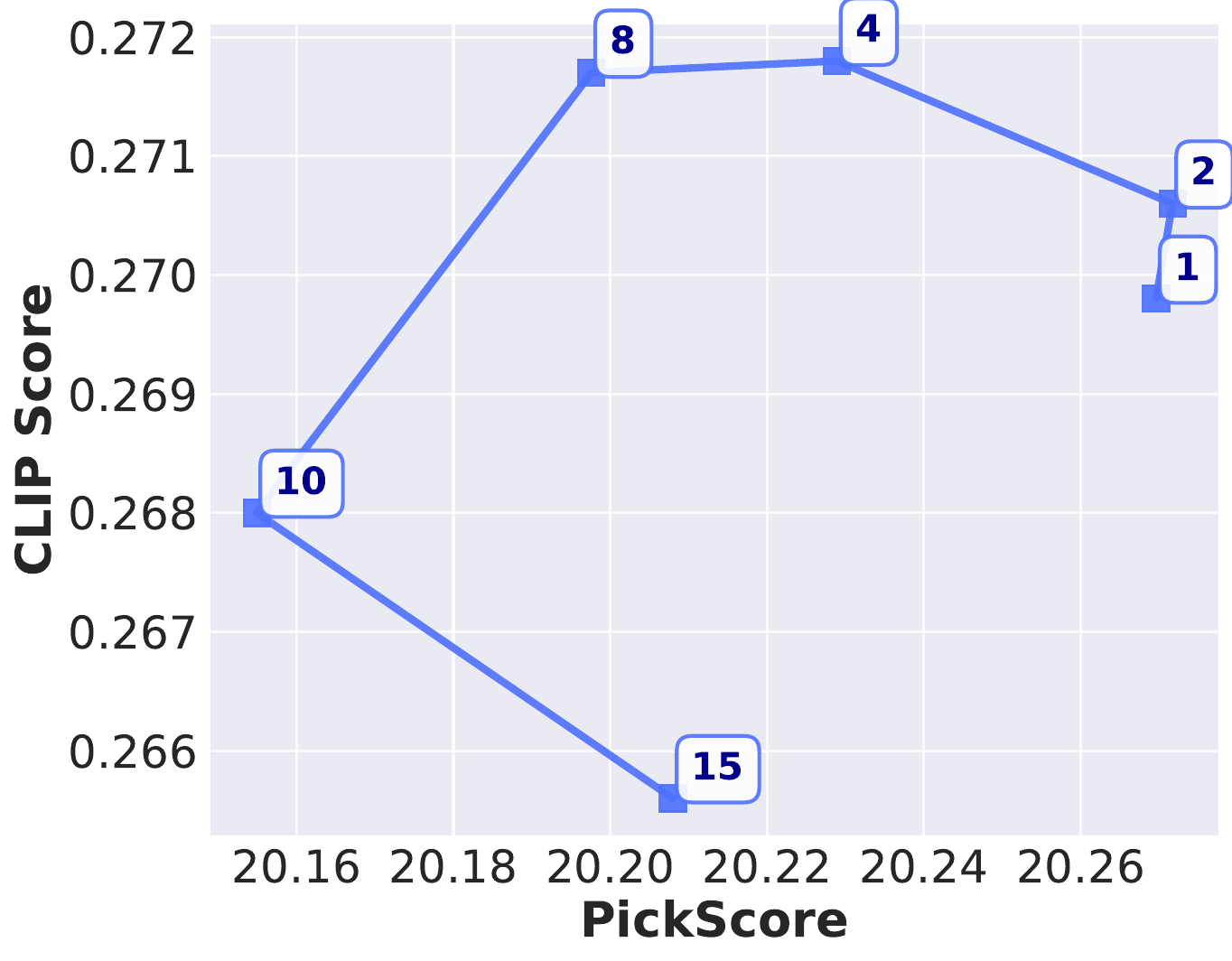}
        \label{fig:pickvsclip}
    \end{subfigure}%
    \hfill
    \begin{subfigure}[t]{0.24\textwidth}
        \centering
        \includegraphics[width=\linewidth]{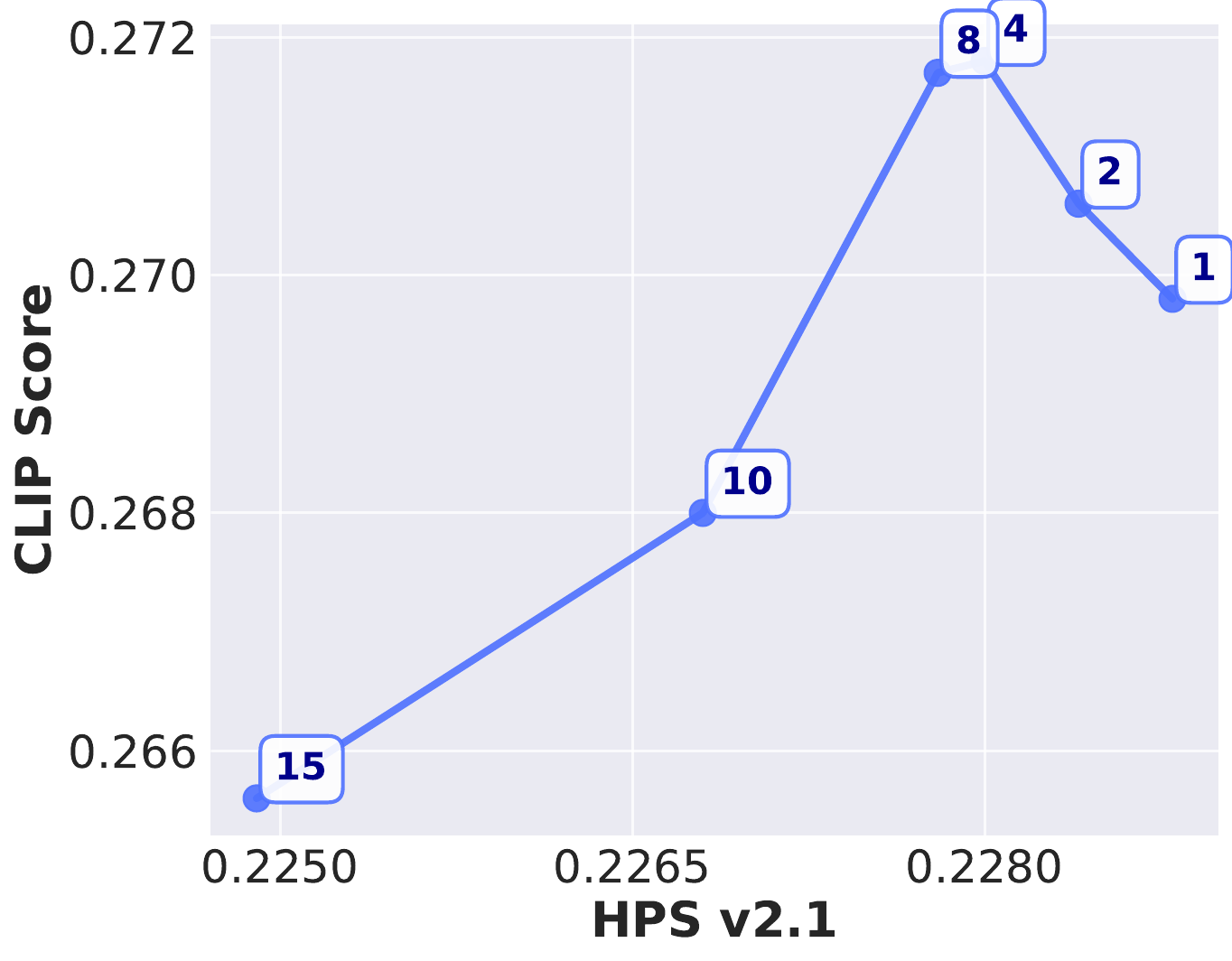}
        \label{fig:hpsvsclip1}
    \end{subfigure}%
    \vspace{-1em}
    \caption{\textbf{Impact of semantic-weighting coefficient $\alpha$ on alignment metrics.} Each point corresponds to a different $\alpha$ value (shown in labels), tracing the trade-off between CLIP Score and human-preference metrics: PickScore (left) and HPS v2.1 (right).}
    \label{fig:hyperparameter}
\end{figure}
%%%%%%%%%%%%%%%%%%%%%%%

\paragraph{Hyperparameter Analysis.}
In practice, $\alpha \approx 4$ is recommended when preserving meaning is paramount, while $\alpha \approx 8$ is preferable when a small semantic trade-off is acceptable for maximising human preference.
Varying the semantic-weighting coefficient $\alpha$ reveals a trade-off between semantic fidelity and preference alignment (Figure~\ref{fig:hyperparameter}). For small values ($\alpha = 1$--$2$), the exponential weight $W_{\alpha} \approx 1$, and the loss behaves like standard DPO, yielding only modest gains in CLIP, HPSv2.1, and PickScore. Increasing $\alpha$ to $4$ sharpens the weighting, filtering out semantically drifting samples while retaining a large effective batch. This yields the best semantic consistency (CLIP $\approx 0.272$) along with strong preference alignment. At $\alpha = 8$, further regularization slightly reduces CLIP ($\approx 0.27$) and slightly reduces PickScore ($\approx 20.26$), suggesting mild semantic penalties can enhance human preference. However, $\alpha \ge 10$ over-suppresses informative gradients, shrinking the batch and reducing all metrics ($\alpha = 15$: CLIP $\approx 0.266$). Correlation tests support this: CLIP and HPSv2.1 are strongly correlated ($r = 0.84$, $p = 0.034$), while CLIP–PickScore shows weaker correlation ($r = 0.35$), explaining why preference peaks at a higher $\alpha$. In practice, $\alpha \approx 4$ is optimal for meaning preservation and maximizing human preference.

\section{Conclusion}
\label{sec:Conc}
DPO maximizes human preference but, because its regularizer is token-level, it leaves \emph{semantic drift} unchecked as preferred prompts can still diverge from the user's intended meaning.
We recast this drift as an explicit optimization objective and introduce \emph{Sem-DPO}, which scales each DPO loss term by an exponential weight proportional to the cosine distance between the original and candidate prompt embeddings, thereby softly down-weighting semantically mismatched pairs.
This semantic-weighted extension preserves DPO's simplicity and efficiency while providing a provable bound on how far optimized prompts may deviate from the input prompt.
\section*{Limitations}
While Sem-DPO offers significant improvements, several limitations warrant consideration for future work. The current implementation relies on a fixed, pre-trained, external frozen embedding model ($e_{\varphi}$) to compute semantic similarity. The choice and robustness of this embedding model can directly influence the effectiveness of the semantic weighting. Exploring adaptive or task-specific embedding functions could potentially further enhance performance. Additionally, the hyperparameter $\alpha$, which controls the strength of the semantic weighting, was set to 8 after manual tuning. While this yielded strong results, the optimal value of $\alpha$ may vary depending on the dataset characteristics or the specific task, suggesting a need for more robust hyperparameter tuning strategies or automated methods for its determination. Finally, while Sem-DPO addresses semantic inconsistency, future research could explore its applicability and performance in addressing other potential issues in prompt optimization, such as stylistic overfitting not directly tied to semantic drift.

\section*{Ethical Considerations}
Semantic-DPO’s efficiency may lower the barrier for malicious or deceptive image generation, so any deployment should couple it with rigorous content-safety filters and provenance tools.
Because the semantic weight inherits biases from CLIP embeddings and human-preference data, practitioners must audit and diversify these sources to avoid reinforcing stereotypes.
Our semantic-drift guarantee rests on cosine distance accurately reflecting meaning, in domains where that proxy fails, human oversight is required to catch residual errors.

\section*{Acknowledgments}
The work of Azal Ahmad Khan was supported in part by the Amazon Machine Learning Systems Fellowship and the UMN GAGE Fellowship.
Xinran Wang and Ali Anwar were supported by the 3M Science and Technology Graduate Fellowship and the Samsung Global Research Outreach Award.

\bibliography{anthology,custom}
\bibliographystyle{acl_natbib}  

\newpage
\clearpage
\onecolumn
\appendix

\section{Appendix}
\label{sec:appendix}

Tables~\ref{table:qwen_results} \& \ref{table:gpt_results} give the complete metric values for every language model and dataset pair.
Table~\ref{table:qwen_results} (Qwen-1.5 B) shows Sem-DPO topping across all three metrics for all the datasets.
Table~\ref{table:gpt_results} (GPT-2) repeats the pattern, with Sem-DPO reclaiming the lead even against specialized prompt-engineering baselines.
These full numbers support the main-text claim that Sem-DPO yields state-of-the-art preference and alignment across models and domains.
Table~\ref{table:human_input_vs_sdpo} shows images generated with human input and semantic DPO.

\definecolor{coco_color}{HTML}{FFE8CD}      % Light blue for COCO
\definecolor{ddb_color}{HTML}{FFF2EB}       % Light green for DDB
\definecolor{lexica_color}{HTML}{FFEAEA}    % Light orange for Lexica

% First table - Qwen methods
\begin{table*}[h]
\centering   
\footnotesize
{%
\begin{tabular}{lcccc}
\noalign{\hrule height 1pt} 
\multicolumn{1}{c|}{\multirow{2}{*}{\small\textbf{Method}}} & \multicolumn{4}{c|}{\small\textit{\textbf{Metrics}}} \\ \cline{2-5} 
\multicolumn{1}{c|}{} & \small\textbf{CLIP Score} & \small\textbf{HPS v2.1} & \small\textbf{PickScore} & \small\textbf{Avg. Score} \\
\hline
\hline
\cellcolor{coco_color}Human Input & \cellcolor{coco_color}0.2599 & \cellcolor{coco_color}0.2572 & \cellcolor{coco_color}21.4127 & \cellcolor{coco_color}0.786 \\
\cellcolor{coco_color}SFT & \cellcolor{coco_color}0.2359 & \cellcolor{coco_color}0.2247 & \cellcolor{coco_color}20.5758 & \cellcolor{coco_color}0.000 \\
\cellcolor{coco_color}DPO & \cellcolor{coco_color}0.2559 & \cellcolor{coco_color}0.2498 & \cellcolor{coco_color}21.0490 & \cellcolor{coco_color}0.572 \\
\cellcolor{coco_color}\textbf{Sem-DPO} & \cellcolor{coco_color}\textbf{0.2618} & \cellcolor{coco_color}\textbf{0.2749} & \cellcolor{coco_color}\textbf{21.6433} & \cellcolor{coco_color}\textbf{1.000} \\
\hline
\cellcolor{ddb_color}Human Input & \cellcolor{ddb_color}0.2600 & \cellcolor{ddb_color}0.2171 & \cellcolor{ddb_color}19.7258 & \cellcolor{ddb_color}0.524 \\
\cellcolor{ddb_color}SFT & \cellcolor{ddb_color}0.2368 & \cellcolor{ddb_color}0.2140 & \cellcolor{ddb_color}19.5800 & \cellcolor{ddb_color}0.000 \\
\cellcolor{ddb_color}DPO & \cellcolor{ddb_color}0.2476 & \cellcolor{ddb_color}0.2162 & \cellcolor{ddb_color}19.6595 & \cellcolor{ddb_color}0.280 \\
\cellcolor{ddb_color}\textbf{Sem-DPO} & \cellcolor{ddb_color}\textbf{0.2629} & \cellcolor{ddb_color}\textbf{0.2314} & \cellcolor{ddb_color}\textbf{19.8464} & \cellcolor{ddb_color}\textbf{1.000} \\
\hline
\cellcolor{lexica_color}Human Input & \cellcolor{lexica_color}0.2679 & \cellcolor{lexica_color}0.2224 & \cellcolor{lexica_color}20.0578 & \cellcolor{lexica_color}0.360 \\
\cellcolor{lexica_color}SFT & \cellcolor{lexica_color}0.2417 & \cellcolor{lexica_color}0.2174 & \cellcolor{lexica_color}20.0973 & \cellcolor{lexica_color}0.058 \\
\cellcolor{lexica_color}DPO & \cellcolor{lexica_color}0.2553 & \cellcolor{lexica_color}0.2232 & \cellcolor{lexica_color}20.1131 & \cellcolor{lexica_color}0.322 \\
\cellcolor{lexica_color}\textbf{Sem-DPO} & \cellcolor{lexica_color}\textbf{0.2734} & \cellcolor{lexica_color}\textbf{0.2371} & \cellcolor{lexica_color}\textbf{20.2857} & \cellcolor{lexica_color}\textbf{1.000} \\
\noalign{\hrule height 1pt}
\end{tabular}
}
\caption{Comparison of prompt optimization methods with the \textbf{qwen-1.5b} model in Text-to-Image Generation across different metrics. Colors represent different datasets: \colorbox{coco_color}{COCO}, \colorbox{ddb_color}{DiffusionDB}, and \colorbox{lexica_color}{Lexica}. The average score is calculated with all scores normalized into the range [0,1]. The highest-performing optimization is bolded.}
\label{table:qwen_results}
\end{table*}

% Second table - GPT methods
% Second table - GPT methods
\begin{table*}[h]
\centering    
\footnotesize
{%
\begin{tabular}{lcccc}
\noalign{\hrule height 1pt} 
\multicolumn{1}{c|}{\multirow{2}{*}{\small\textbf{Method}}} & \multicolumn{4}{c|}{\small\textit{\textbf{Metrics}}} \\ \cline{2-5}
\multicolumn{1}{c|}{} & \small\textbf{CLIP Score} & \small\textbf{HPS v2.1} & \small\textbf{PickScore} & \small\textbf{Avg. Score} \\
\hline
\hline
\cellcolor{coco_color}Human Input & \cellcolor{coco_color}0.2599 & \cellcolor{coco_color}0.2572 & \cellcolor{coco_color}21.4127 & \cellcolor{coco_color}0.950 \\
\cellcolor{coco_color}SFT & \cellcolor{coco_color}0.2279 & \cellcolor{coco_color}0.2200 & \cellcolor{coco_color}20.3911 & \cellcolor{coco_color}0.000 \\
\cellcolor{coco_color}Beautiful Prompt & \cellcolor{coco_color}0.2287 & \cellcolor{coco_color}0.2275 & \cellcolor{coco_color}20.4893 & \cellcolor{coco_color}0.105 \\
\cellcolor{coco_color}Promptist & \cellcolor{coco_color}0.2512 & \cellcolor{coco_color}0.2372 & \cellcolor{coco_color}21.0774 & \cellcolor{coco_color}0.587 \\
\cellcolor{coco_color}DPO & \cellcolor{coco_color}0.2403 & \cellcolor{coco_color}0.2306 & \cellcolor{coco_color}20.5972 & \cellcolor{coco_color}0.274 \\
\cellcolor{coco_color}\textbf{Sem-DPO} & \cellcolor{coco_color}\textbf{0.2639} & \cellcolor{coco_color}\textbf{0.2581} & \cellcolor{coco_color}\textbf{21.4289} & \cellcolor{coco_color}\textbf{1.000} \\
\hline
\cellcolor{ddb_color}Human Input & \cellcolor{ddb_color}\textbf{0.2600} & \cellcolor{ddb_color}0.2171 & \cellcolor{ddb_color}19.7258 & \cellcolor{ddb_color}0.867 \\
\cellcolor{ddb_color}SFT & \cellcolor{ddb_color}0.2080 & \cellcolor{ddb_color}0.2035 & \cellcolor{ddb_color}19.2212 & \cellcolor{ddb_color}0.000 \\
\cellcolor{ddb_color}Beautiful Prompt & \cellcolor{ddb_color}0.2175 & \cellcolor{ddb_color}\textbf{0.2210} & \cellcolor{ddb_color}19.6115 & \cellcolor{ddb_color}0.608 \\
\cellcolor{ddb_color}Promptist & \cellcolor{ddb_color}0.2436 & \cellcolor{ddb_color}0.2192 & \cellcolor{ddb_color}\textbf{19.8338} & \cellcolor{ddb_color}\textbf{0.866} \\
\cellcolor{ddb_color}DPO & \cellcolor{ddb_color}0.2427 & \cellcolor{ddb_color}0.2171 & \cellcolor{ddb_color}19.6598 & \cellcolor{ddb_color}0.725 \\
\cellcolor{ddb_color}\textbf{Sem-DPO} & \cellcolor{ddb_color}0.2531 & \cellcolor{ddb_color}0.2181 & \cellcolor{ddb_color}19.7519 & \cellcolor{ddb_color}0.862 \\
\hline
\cellcolor{lexica_color}Human Input & \cellcolor{lexica_color}\textbf{0.2679} & \cellcolor{lexica_color}0.2224 & \cellcolor{lexica_color}20.0578 & \cellcolor{lexica_color}0.841 \\
\cellcolor{lexica_color}SFT & \cellcolor{lexica_color}0.2213 & \cellcolor{lexica_color}0.2081 & \cellcolor{lexica_color}19.7741 & \cellcolor{lexica_color}0.000 \\
\cellcolor{lexica_color}Beautiful Prompt & \cellcolor{lexica_color}0.2280 & \cellcolor{lexica_color}0.2240 & \cellcolor{lexica_color}20.0134 & \cellcolor{lexica_color}0.553 \\
\cellcolor{lexica_color}Promptist & \cellcolor{lexica_color}0.2528 & \cellcolor{lexica_color}0.2230 & \cellcolor{lexica_color}\textbf{20.2028} & \cellcolor{lexica_color}0.858 \\
\cellcolor{lexica_color}DPO & \cellcolor{lexica_color}0.2417 & \cellcolor{lexica_color}0.2161 & \cellcolor{lexica_color}20.0635 & \cellcolor{lexica_color}0.532 \\
\cellcolor{lexica_color}\textbf{Sem-DPO} & \cellcolor{lexica_color}0.2604 & \cellcolor{lexica_color}\textbf{0.2247} & \cellcolor{lexica_color}20.1341 & \cellcolor{lexica_color}\textbf{0.893} \\
\noalign{\hrule height 1pt}
\end{tabular}
}
\caption{Comparison of prompt optimization methods with the \textbf{GPT-2} model in Text-to-Image Generation across different metrics. Colors represent different datasets: \colorbox{coco_color}{COCO}, \colorbox{ddb_color}{DiffusionDB}, and \colorbox{lexica_color}{Lexica}. The average score is calculated with all scores normalized into the range [0,1]. The highest-performing optimization is bolded.}
\label{table:gpt_results}
\end{table*}

\begin{table}[!htbp]
  \centering
  \small
  \resizebox{\columnwidth}{!}{%
    \begin{tabular}{p{2.5cm}p{2.5cm}}
      \toprule
      \multicolumn{1}{c}{\textbf{Human Input}}
        & \multicolumn{1}{c}{\textbf{Sem-DPO}} \\
      \midrule
      \multicolumn{1}{c}{\raisebox{-0.5\height}{\includegraphics[width=2.1cm,height=2.1cm]{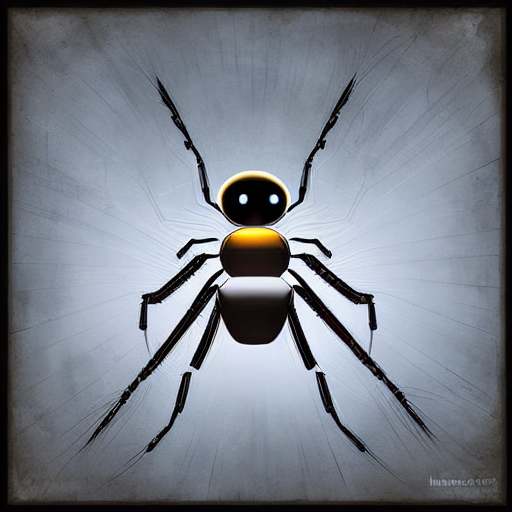}}}
      & \multicolumn{1}{c}{\raisebox{-0.5\height}{\includegraphics[width=2.1cm,height=2.1cm]{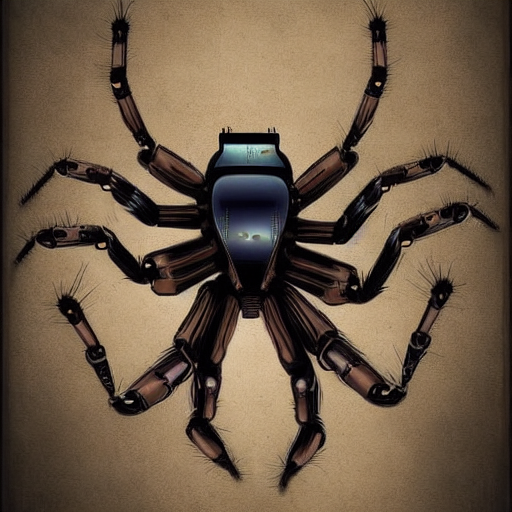}}} \\
      \multicolumn{1}{c}{Prompt 1: “Realistic robotic spider poster”}
      & \multicolumn{1}{c}{Prompt 2: “Realistic robotic spider poster, highly detailed, digital art by.”} \\
      \midrule
      \multicolumn{1}{c}{\raisebox{-0.5\height}{\includegraphics[width=2.1cm,height=2.1cm]{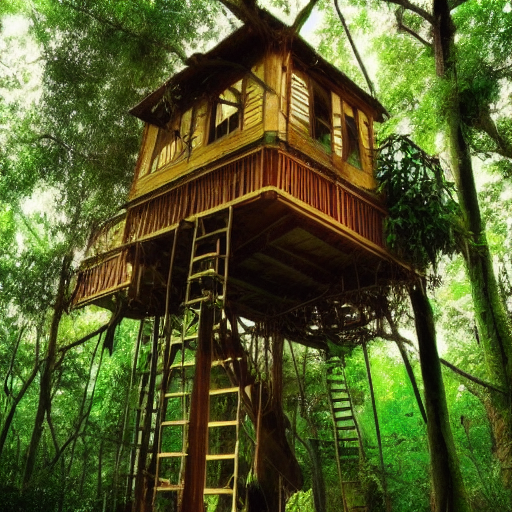}}}
      & \multicolumn{1}{c}{\raisebox{-0.5\height}{\includegraphics[width=2.1cm,height=2.1cm]{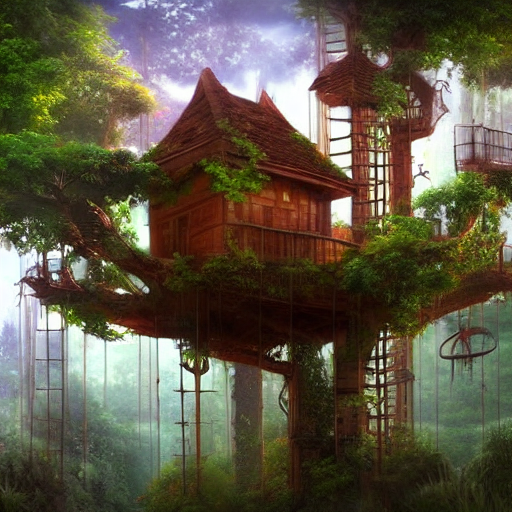}}} \\
      \multicolumn{1}{c}{Prompt 3: “Cinematic beautiful jungle tree house”}
      & \multicolumn{1}{c}{Prompt 4: “Cinematic beautiful jungle tree house, very detailed and fantasy landscape”} \\
      \bottomrule
    \end{tabular}%
  }
  \caption{Examples of images generated using human input and Sem-DPO prompts.}
  \label{table:human_input_vs_sdpo}
\end{table}

\newpage

Table~\ref{table:human_input_vs_sdpo} illustrates qualitative differences in image generations using original human-written prompts versus Sem-DPO optimized prompts, showcasing enhanced detail and stylistic richness in the latter.

% \section{Example Appendix}
% \label{sec:appendix}

% This is a section in the appendix.

\end{document}